\documentclass[sigconf]{acmart}
\AtBeginDocument{%
  }

\usepackage{algorithmic}
\usepackage{graphicx}
\usepackage{subfigure}
\usepackage{makecell}
\usepackage{subcaption}
\usepackage{multirow}
\usepackage{colortbl}
\usepackage{algorithm}
\usepackage{esvect}
\usepackage{makecell}
\usepackage{microtype}
\newcommand{\method}{DisenMamba}
\newcommand{\task}{NTAD}

\copyrightyear{2026}
\acmYear{2026}
\setcopyright{cc}
\setcctype{by}
\acmConference[KDD '26]{Proceedings of the 32nd ACM SIGKDD Conference on Knowledge Discovery and Data Mining V.2}{August 09--13, 2026}{Jeju Island, Republic of Korea}
\acmBooktitle{Proceedings of the 32nd ACM SIGKDD Conference on Knowledge Discovery and Data Mining V.2 (KDD '26), August 09--13, 2026, Jeju Island, Republic of Korea}
\acmDOI{10.1145/3770855.3817980}
\acmISBN{979-8-4007-2259-2/2026/08}

\begin{document}


\title{Disentangling Multi-View Scanning in Mamba for Network Traffic Anomaly Detection}

\author{Xinglin Lian}
\orcid{0009-0000-0627-8933}
\email{kenshin.lian24@gmail.com}
\affiliation{%
  \institution{University of Electronic Science and Technology of China}
  \city{Chengdu}
  \state{Sichuan}
  \country{China}
}

\author{Chengtai Cao}
\orcid{0000-0003-3944-8358}
\authornote{Corresponding Authors.}
\email{chengtcao2-c@my.cityu.edu.hk}
\affiliation{%
  \institution{University of Electronic Science and Technology of China}
  \city{Chengdu}
  \state{Sichuan}
  \country{China}
}

\affiliation{%
\institution{City University of Hong Kong}
  \city{Hong Kong SAR}
  \country{China}
}

\author{Ting Zhong}
\orcid{0000-0002-8163-3146}
\email{zhongting@uestc.edu.cn}
\affiliation{%
  \institution{University of Electronic Science and Technology of China}
  \city{Chengdu}
  \state{Sichuan}
  \country{China}
}

\author{Fan Zhou}
\orcid{0000-0002-8038-8150}
\authornotemark[1]
\email{fan.zhou@uestc.edu.cn}
\affiliation{%
  \institution{University of Electronic Science and Technology of China}
  \city{Chengdu}
  \state{Sichuan}
  \country{China}
}

  \affiliation{%
\institution{Key Laboratory of Intelligent Digital Media Technology of Sichuan Province}
  \city{Chengdu}
  \state{Sichuan}
  \country{China}
}

\renewcommand{\shortauthors}{Xinglin Lian, Chengtai Cao, Ting Zhong, and Fan Zhou}

\begin{abstract}
    Network Traffic Anomaly Detection (\task) is a critical task in cybersecurity, yet timely and accurate anomaly detection remains challenging. Mamba has emerged as a particularly promising backbone for \task~due to its linear-time complexity for long-sequence modeling. It further incorporates a dedicated multi-view scanning mechanism to enhance detection precision through complementary contextual cues. However, we identify a previously overlooked structural deficiency in multi-view Mamba scanning for \task: redundancy accumulation. Specifically, distinct scanning branches capture substantial view-invariant information, which is repeatedly amplified during multi-view fusion; conversely, view-specific information is diluted or even suppressed, leading to representation homogenization and multi-view degradation. To address this problem, we propose \textbf{\method}, a novel disentangled multi-view Mamba framework. \method~reformulates multi-view scanning as a two-stage disentangle-then-fuse process that explicitly separates view-invariant and view-specific components prior to fusion. This design prevents the invariant information accumulation while preserving complementary multi-view cues, yielding more discriminative representations for subtle traffic anomalies. Extensive experiments demonstrate the effectiveness of~\method, establishing a new disentangled multi-view Mamba paradigm. Code is available at \url{https://github.com/ikun0124/DisenMamba}.
\end{abstract}




\maketitle


\section{Introduction}
\label{sec:Introduction}
Network Traffic Anomaly Detection (\task) is a fundamental task in modern cybersecurity, aiming to monitor real-time network traffic and identify unauthorized or malicious connections~\cite{yang2024recda,zhong2026shattering}. Effective anomaly detection plays a critical role in ensuring high-quality service availability, protected user privacy, and reliable data security~\cite{zhang2025revolutionizing,zhong2025noise}. Despite its importance, the widespread adoption of payload encryption and the exponential growth of traffic volume make accurate and timely anomaly detection increasingly challenging~\cite{wickramasinghe2025sok, yang2025mm4flow}. For example, signature-matching approaches have become largely ineffective, revealing the fundamental limitations of traditional rule-based solutions~\cite{zheng2024multi,he2026hierarchical}.

Benefiting from strong representation learning capabilities, deep learning-based methods have become the dominant approach for \task. According to network architectures, existing approaches can be broadly categorized into multilayer perceptrons (MLPs)~\cite{zong2018deep, yang2025unflows}, convolutional neural networks (CNNs)~\cite{lunardi2023arcade, lian2024payload}, recurrent neural networks (RNNs)~\cite{miao2024boosting, westphal2025feature}, and Transformers~\cite{chen2025miett, zhao2025towards}. Despite continuous progress, these architectures still exhibit inherent limitations for \task. Specifically, MLP- and CNN-based methods struggle to capture multi-view and long-range dependencies, where anomalous traffic patterns are often embedded in flow-level and packet-level interactions~\cite{lian2026contextual,zhao2024novel}. Transformers incur prohibitive quadratic complexity, while RNN-based architectures suffer from limited parallelism, both of which hinder real-time deployment. Recently, State Space Model (SSM)–based Mamba enables efficient long-sequence modeling with linear-time complexity via selective state-space mechanisms~\cite{gu2024mamba}. Combined with its multi-view scanning design for complementary contextual modeling, Mamba becomes a particularly promising backbone for \task.

\textbf{Challenges.} The multi-view scanning mechanism is a core and indispensable design in Mamba, enabling complementary contextual modeling while alleviating long-term dependency decay~\cite{liu2025vision, gao2025ssd}. Nevertheless, recent studies on network traffic analysis show that naively applying this mechanism can unexpectedly degrade detection performance~\cite{wang2024netmamba}. To avoid such degradation, subsequent works adopt simplified unidirectional Mamba architectures~\cite{deng2025countmamba, xia2025mamba4net}. However, unidirectional scanning inherently introduces causal bias and limits global contextual modeling~\cite{qu2024survey}, which contradicts the multi-view nature of network traffic. Consequently, existing Mamba-based methods merely benefit from computational efficiency, while underutilizing their multi-view contextual modeling potential. This raises a critical question: \textbf{Is Mamba inherently unsuitable for \task, or is the performance bottleneck caused by suboptimal multi-view scanning designs?}

\begin{figure}[t]
\centering
\includegraphics[width=\linewidth]{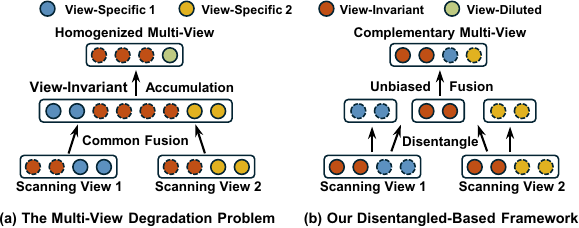}
\vspace{-0.55cm}
\caption{(a) Redundancy accumulation in conventional multi-view Mamba scanning. (b) The proposed disentangle-then-fuse scanning framework for mitigating this problem.}
\label{fig:Motivation}
\vspace{-0.25cm}
\end{figure}

\textbf{Key Finding.} To answer this question, we conduct extensive empirical investigations and uncover a previously overlooked structural deficiency in multi-view scanning. This deficiency manifests as \textbf{redundancy accumulation}, which constitutes a fundamental performance bottleneck for Mamba-based \task~models. As illustrated in~\figurename~\ref{fig:Motivation}(a), multi-view branches inevitably capture substantial inter-view invariant information. During conventional view fusion, these invariant components increasingly dominate the fused representation and are further amplified as more scanning branches are introduced. In contrast, view-specific information is gradually diluted and may even be suppressed, leading to the loss of directional semantics. As a result, the fused representations become progressively homogenized toward invariant patterns, causing multi-view Mamba architectures to perform even worse than their single-view counterparts. We formally define this effect as the \textbf{multi-view degradation problem} and identify it as a key obstacle to fully unlocking Mamba’s potential for \task.

\textbf{Our Solution.} To address the multi-view degradation problem, we propose \textbf{\method}, a novel Disentangled Mamba framework for \task~task. Considering the inherent multi-view nature of network traffic, we first design a multi-view scanning strategy along forward–backward flow-level and packet-level dimensions to capture complementary traffic characteristics. As shown in~\figurename~\ref{fig:Motivation}(b), we reformulate multi-view scanning as a two-stage process: (i) explicitly disentangling view-invariant and view-specific components, followed by (ii) fusing the disentangled component-level representations. This disentangle-then-fuse design balances the contributions of invariant and specific components, thereby preventing inter-view redundancy accumulation while preserving complementary multi-view information. As a result, the learned representations become more sensitive to subtle anomalous traffic patterns. To realize this design, we employ lightweight projection modules to extract view-invariant and view-specific information, and introduce a disentanglement constraint loss that facilitates their separation during training. The proposed framework introduces only negligible inference overhead compared to original Mamba-based models, while remaining easy to implement and deploy. \textbf{In summary, our contributions are fourfold:}
\begin{itemize}
\item We are the first to systematically identify a redundancy accumulation phenomenon and the resulting multi-view degradation problem in Mamba-based models for \task.
\item We propose a multi-view scanning strategy and a novel disentangled Mamba framework that explicitly separates view-invariant and view-specific information, effectively alleviating multi-view representation degradation.
\item We introduce lightweight projection modules and a disentanglement constraint loss for efficient multi-view Mamba representation disentanglement.
\item Extensive experiments on three real-world datasets show that \method~outperforms state-of-the-art NTAD methods in both effectiveness and efficiency.
\end{itemize}
\section{Motivation}\label{sec:Motivation}
\subsection{Multi-View Scanning}
Multi-view scanning has been established as an essential component of Mamba-based models, with widespread adoption in computer vision~\cite{liu2025vision, huang2024localmamba} and time-series analysis~\cite{gao2025ssd, Wang2025ismamba}. This design is originally motivated by the inherent limitations of primitive unidirectional Mamba. Specifically, unidirectional selective scanning introduces causal bias and restricts global contextual modeling~\cite{qu2024survey}. Moreover, its ability to preserve long-range dependencies is limited, which can lead to catastrophic forgetting under extremely long contexts~\cite{wang2025understanding}. Yet, recent studies show that naive multi-view scanning degrades detection performance on network traffic, even underperforming single-view scanning~\cite{wang2024netmamba}. In response, subsequent works adopt simplified unidirectional Mamba architectures~\cite{deng2025countmamba, xia2025mamba4net}. However, unidirectional scanning fails to capture anomalous traffic patterns inherently embedded in multi-view interactions across flow-level and packet-level contexts~\cite{lian2026contextual, zhao2024novel}. This counterintuitive observation motivates a deeper investigation into the representational behavior of Mamba for \task.

\subsection{Redundancy Accumulation Phenomenon}\label{sec:Redundancy Accumulation Phenomenon}
\textbf{Experimental Setup.} To this end, we quantitatively analyze representation dynamics during multi-view fusion using two representative Mamba-based methods, SMamba~\cite{Wang2025ismamba} and MambaAD~\cite{qin2025mambaad}, on three real-world network traffic datasets: CIC-IoT2023~\cite{neto2023ciciot2023}, DoHBrw2020~\cite{montazerishatoori2020detection}, and ISCX-Tor2016~\cite{lashkari2017characterization}. Specifically, SMamba employs two non-shared Mamba layers for bidirectional scanning, whereas MambaAD adopts a parameter-shared bidirectional design for multi-view modeling. We measure inter-view relationships before and after fusion using linear centered kernel alignment (CKA) similarity, a widely used metric for quantifying subspace similarity and representational overlap~\cite{tran2025multi,kornblith2019similarity,lappe2025register}.
\begin{equation}
\mathrm{CKA}(\hat {\boldsymbol H_1}, \hat{\boldsymbol H_2})
=
\frac{
\left\|  \hat{\boldsymbol H}_1^{\top} \hat{\boldsymbol H}_2 \right\|_{Fro}^2
}{
\left\| \hat{\boldsymbol H}_1^{\top} \hat{\boldsymbol H}_1 \right\|_{Fro}
\left\| \hat{\boldsymbol H}_2^{\top} \hat{\boldsymbol H}_2 \right\|_{Fro}
},
\end{equation}
where $\hat{\boldsymbol H}_1$ and $\hat{\boldsymbol H}_2$ denote the two feature matrices after decentralization, and $||\cdot||_{Fro}$ denotes the Frobenius norm.

\begin{figure}[t]
\centering
\begin{minipage}{0.50\columnwidth}
  \centering
  \includegraphics[width=\linewidth]{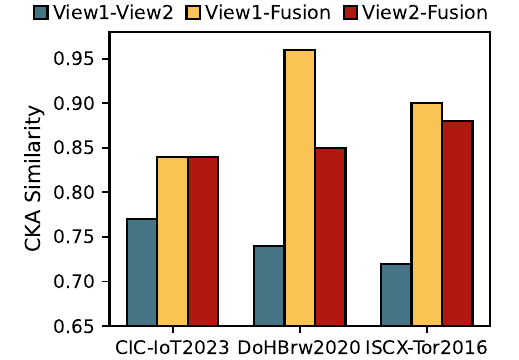}
  \vspace{-0.6cm}
  \caption*{(a) SMamba}
\end{minipage}\hfill
\begin{minipage}{0.50\columnwidth}
  \centering
  \includegraphics[width=\linewidth]{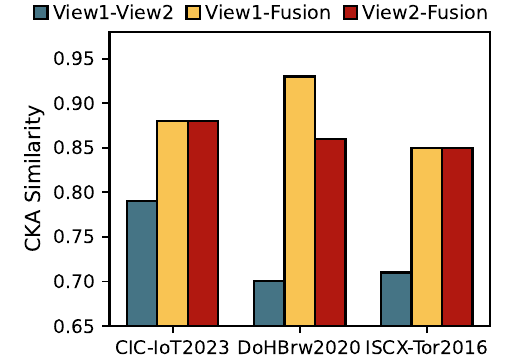}
  \vspace{-0.6cm}
  \caption*{(b) MambaAD}
\end{minipage}
\vspace{-0.3cm}
\caption{Inter-view redundancy statistics for baselines.}
\label{fig:redundancy1}
\vspace{-0.2cm}
\end{figure}

\begin{figure}[t]
\centerline{\includegraphics[width=0.9\linewidth]{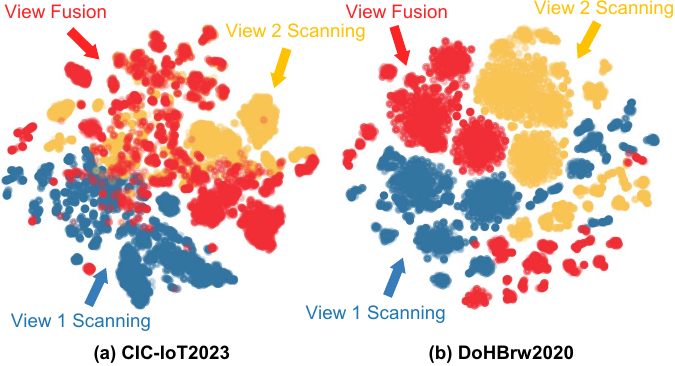}}
\vspace{-0.3cm}
\caption{t-SNE visualization of learned representations across different scanning views.}
\label{fig:t-sne1}
\vspace{-0.2cm}
\end{figure}

\textbf{Multi-View Degradation Problem.}
As shown in~\figurename~\ref{fig:redundancy1}, both SMamba and MambaAD exhibit consistently high inter-view similarity (View1--View2), indicating that distinct scanning branches capture highly overlapping representations dominated by view-invariant information. We further visualize the learned features using t-distributed Stochastic Neighbor Embedding (t-SNE) in~\figurename~\ref{fig:t-sne1}. The highly symmetric inter-view distributions suggest that multi-view branches primarily encode view-invariant patterns rather than view-specific characteristics. The fused representations further remain similar to the individual views, indicating that invariant components are repeatedly reinforced during fusion, resulting in \textbf{redundancy accumulation}. As evidenced in~\figurename~\ref{fig:redundancy1}, the fused representations exhibit substantially higher similarities to individual views (View1--Fusion and View2--Fusion) than the original inter-view similarity (View1--View2). This observation indicates that fusion introduces little genuinely view-specific information; instead, dominant invariant components are further amplified (if cross-view specific information were incorporated, fused similarity would decrease). Consequently, the fused representations become increasingly homogenized, progressively diluting the marginal view-specific signals. As illustrated in~\figurename~\ref{fig:Motivation}(a), each scanning branch gradually loses its original directional characteristics. We formally define this as the \textbf{multi-view degradation problem}. Ultimately, this leads to a counterintuitive outcome reported in prior studies~\cite{wang2024netmamba}: introducing multi-view scanning can paradoxically underperform single-view scanning.

\subsection{Conflict and Motivation}
\textbf{Mechanism-Level Conflict.}
Mamba relies on selective scanning to extract salient contextual information. In principle, multi-view scanning is intended to capture complementary representations by leveraging selective scanning across different views. However, existing methods typically fuse multi-view features via simple summation or concatenation operations~\cite{qu2024survey, liu2025vision, Wang2025ismamba}. Such designs implicitly encourage scanning branches to repeatedly attend to dominant invariant patterns rather than discriminative view-specific cues. As evidenced by our empirical results, even when independently parameterized Mamba layers are used to model different views (e.g., SMamba~\cite{Wang2025ismamba}), this mechanism-level conflict between selectivity and multi-view fusion remains unresolved. Such representations become view-invariant and homogenized, failing to capture the heterogeneous multi-view anomalies of network traffic~\cite{lian2026contextual}.

\textbf{Motivation.}
To address this issue, we propose a feature-disentangled Mamba architecture. Specifically, we reformulate multi-view scanning as a two-stage process: first, disentangling view-invariant and view-specific components, and then fusing the disentangled representations. Compared with existing fusion strategies, this disentangle-then-fuse design effectively mitigates invariant information accumulation. By facilitating complementary multi-view representation learning, the proposed approach improves sensitivity to subtle anomalous patterns in network traffic, thereby unlocking Mamba’s potential for \task.

\section{Methodology}
\label{sec:Methods}
\begin{figure*}[t]
\centerline{\includegraphics{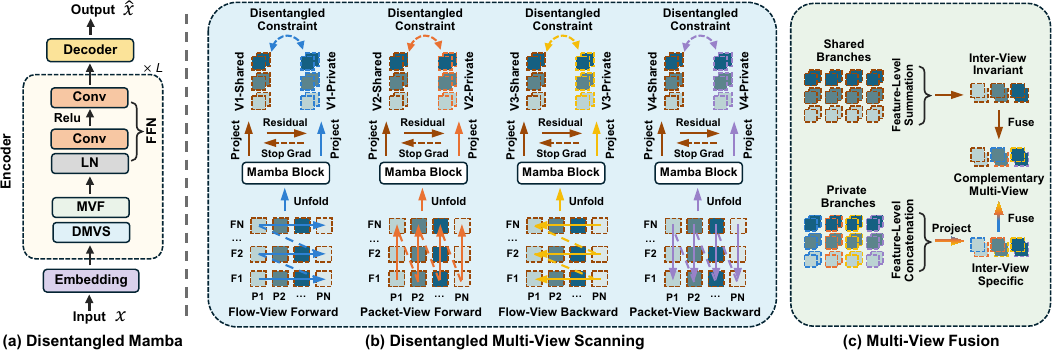}}
\vspace{-0.3cm}
\caption{
(a) {Overview of \method}.
(b) {Disentangled Multi-View Scanning (DMVS module)}: Decomposing each scanning view into shared and private branches.
(c) {Multi-View Fusion (MVF module)}: Integrating disentangled representations to form complementary multi-view features while preventing redundancy accumulation.}
\label{fig:Framework}
\vspace{-0.05cm}
\end{figure*}

\subsection{Problem Definition}
This work addresses the Network Traffic Anomaly Detection (NTAD) task under the widely adopted zero-positive learning setting, where only normal samples are available for training. Compared with traditional supervised learning settings, this paradigm enables effective detection of previously unseen anomalies by measuring deviations from learned normal patterns~\cite{lian2025facing, cai2026scope}. For each test sample, the model outputs an anomaly score $s$, where a larger value indicates a higher likelihood of anomalous activity. Let $\mathcal{X} = \{\boldsymbol{x}_{1}, \boldsymbol{x}_{2},...,\boldsymbol{x}_{N}\}$ denote a training set consisting of $N$ normal samples. Each sample $\boldsymbol{x}$ represents a network flow, defined as a sequence of packets exchanged between endpoints within a specific time window or connection session. We construct each flow sample by segmenting raw traffic into flows, anonymizing address fields, normalizing header lengths, and applying sequence padding/truncation. To capture multi-view traffic characteristics, each flow is formulated as a hierarchical matrix $\boldsymbol{x} \in \mathbb{R}^{F \times P}$, where $F$ denotes the number of packets and $P$ represents the packet size in bytes. More preprocessing details are provided in Appendix~\ref{sec:data preprocessing}.

\subsection{Disentangled Multi-View Scanning Mamba}
This section presents the proposed \method, whose overall architecture is illustrated in \figurename~\ref{fig:Framework}(a). \method~follows an autoencoder-based paradigm, where the input traffic sequence is processed by an embedding module, an encoder, and a decoder to reconstruct the original input. The reconstruction error is used as the anomaly score $s$. The encoder is the core component of \method~and consists of stacked disentangled Mamba blocks and standard feed-forward network (FFN) layers~\cite{xu2022anomaly,cheng2024retrieval}, as shown in \figurename~\ref{fig:Framework}(b) and (c). Each disentangled Mamba block contains two key modules: Disentangled Multi-View Scanning (DMVS), which performs bidirectional flow-level and packet-level scanning while separating view-invariant and view-specific information, and Multi-View Fusion (MVF), which fuses the disentangled representations layer by layer to produce encoded features.

\subsubsection{\textbf{Traffic Embedding}}
Embedding is a fundamental operation for preserving sequential structure and facilitating downstream modeling~\cite{xie2025de3s,chen2026unveiling,chen2025semantic}. Each traffic sample $\boldsymbol{x}$ is tokenized at the byte level, and each byte token is mapped to a $D$-dimensional embedding using two complementary components: a data embedding generated by a $\operatorname{Conv1D}$ layer and a positional embedding produced by a $\operatorname{Sine\text{-}Cosine}$ function to encode sequential order~\cite{xu2022anomaly}. The resulting token representation is defined as:
\begin{equation}
    \boldsymbol{h^{0}} = \operatorname{Conv1D}(\boldsymbol{x}) + \operatorname{Sine-Cosine}(\boldsymbol{x}),
\label{eq:eq1}
\end{equation}
where $\boldsymbol{h^{0}} \in \mathbb{R}^{F \times P \times D}$ encodes the hierarchical information of $\boldsymbol{x}$.

\subsubsection{\textbf{Multi-View Scanning}}
The scanning strategy is a core component of Mamba-based models and has been extensively studied, particularly in the computer vision domain~\cite{liu2025vision, huang2024localmamba}. However, these approaches primarily leverage geometric, i.e., non-adjacent spatial properties, which are not directly applicable to network traffic data. Considering the inherent flow–packet structure of network traffic, we design a task-specific multi-view scanning strategy, as illustrated in \figurename~\ref{fig:Framework}(b). Given the $l$-th layer representation $\boldsymbol{h}^l$, we rearrange the sequence along the flow and packet dimensions to construct forward flow-view and packet-view scanning sequences, denoted as $\smash{{\boldsymbol{h}}^{l}_{\vec{\mathrm{F}}}}$ and $\smash{{\boldsymbol{h}}^{l}_{\vec{\mathrm{P}}}}$, respectively. However, relying solely on forward scanning is insufficient, as unidirectional scanning introduces causal bias and limits global contextual modeling~\cite{qu2024survey}. To mitigate this issue, we further perform backward scanning for both views, denoted as $\smash{{\boldsymbol{h}}^{l}_{\lvec{\mathrm{F}}}}$ and $\smash{{\boldsymbol{h}}^{l}_{\lvec{\mathrm{P}}}}$. The resulting forward and backward representations are then passed to standard Mamba blocks~\cite{gu2024mamba} to encode flow-level and packet-level multi-view dependencies. Compared with existing Mamba-based methods for network traffic analysis~\cite{wang2024netmamba, deng2025countmamba}, this design enables complementary multi-view modeling and more effective exposure of anomalous patterns.
\begin{equation}
\boldsymbol{\tilde h}^{\,l}_{v}
= \operatorname{MambaBlock}\!\left(\boldsymbol{h}^{\,l}_{v}
\right),
\quad
v \in \mathcal{V} = \{\vec{\mathrm{F}}, \vec{\mathrm{P}}, \lvec{\mathrm{F}}, \lvec{\mathrm{P}}\},
\label{eq:eq8}
\end{equation}
where $\boldsymbol{\tilde h}^{l}_{v}$ is the Mamba-encoded representation of view $v$, and $v$ indexes one scanning view in $\mathcal{V}=\{\vec{\mathrm{F}}, \vec{\mathrm{P}}, \lvec{\mathrm{F}}, \lvec{\mathrm{P}}\}$.

\subsubsection{\textbf{Inter-View Disentanglement}}
Existing approaches typically integrate these features using naive summation or concatenation operations~\cite{liu2025vision, Wang2025ismamba}. As revealed in motivation analysis (Section \ref{sec:Motivation}), such fusion strategies overlook inter-view redundancy, leading to redundancy accumulation that suppresses informative view-specific cues and may even degrade detection performance. To address this problem, we disentangle view-invariant and view-specific components before fusion. Specifically, we propose a disentangled Mamba architecture with explicit structural inter-view disentanglement. As illustrated in \figurename~\ref{fig:Framework}(b), each scanning view is decomposed into a shared branch and a private branch. The shared branch captures view-invariant information (i.e., common patterns across views), whereas the private branch preserves view-specific information (i.e., complementary directional cues).

As discussed in Section~\ref{sec:Redundancy Accumulation Phenomenon}, view-invariant information dominates across scanning views and naturally tends to aggregate during fusion. Accordingly, we employ a parameter-shared projection matrix to extract inter-view shared representations:
\begin{equation}
\boldsymbol{\tilde h}^{\,l}_{v,s}
= \boldsymbol{\tilde h}^{\,l}_{v}\,\mathbf{W}^{l}_{s},
\label{eq:shared}
\end{equation}
where $\mathbf{W}^{l}_{s}\in\mathbb{R}^{D\times D}$ denotes the shared projection matrix.

In contrast, private branches are designed to capture incremental view-specific information. Since view-specific components are substantially weaker than view-invariant ones, directly projecting private features is prone to severe interference from dominant shared patterns. To alleviate this interference, we introduce a residual-based extraction mechanism, as illustrated in \figurename~\ref{fig:Framework}(b). Specifically, view-invariant components are first removed from each scanning representation, and the resulting residuals are then passed through independently parameterized projections to extract private features. Notably, we adopt branch-specific low-rank projections for the private branches. This design aligns with the sparse nature of view-specific information. It constrains information throughput and prevents unnecessary shared components from leaking into private branches. Our experiments further demonstrate that this operation is critical for effective feature disentanglement. The private branch is formulated as follows:
\begin{equation}
\boldsymbol{\tilde h}^{\,l}_{v,p}
=
\bigl(
\boldsymbol{\tilde h}^{\,l}_{v}
-
\operatorname{StopGrad}(
\boldsymbol{\tilde h}^{\,l}_{v,s}
)
\bigr)
\mathbf{W}^{l}_{v,p,1}\mathbf{W}^{l}_{v,p,2},
\label{eq:private}
\end{equation}
where $\mathbf{W}^{l}_{v,p,1}\in\mathbb{R}^{D\times R}$ and $\mathbf{W}^{l}_{v,p,2}\in\mathbb{R}^{R\times D}$ form a low-rank factorization with $R\ll D$ for the $v$-th private branch, and $\operatorname{StopGrad}(\cdot)$ prevents gradient propagation from the shared branch during private representation extraction process.

\subsubsection{\textbf{Multi-View Fusion}}
The core of this work lies in integrating multi-view scanning with explicit feature disentanglement. Unlike existing Mamba-based methods that fuse multi-view representations in an indiscriminate manner, our disentangle-then-fuse paradigm enforces balanced contributions from view-invariant and view-specific components through an explicit architectural design. This design avoids representation homogenization and enables effective multi-view modeling. By preventing repeated accumulation of view-invariant components while preserving complementary multi-view information, our approach unlocks the representational capacity of multi-view Mamba.

The disentangled fusion process is illustrated in \figurename~\ref{fig:Framework}(c). For the shared branches, we apply feature-level summation to aggregate inter-view invariant representations. Since shared features are extracted using parameter-shared projection matrices, this summation further reinforces and refines view-invariant information, which naturally dominates across scanning views. For the private branches, we concatenate view-specific representations and apply a projection matrix to integrate complementary inter-view information. Finally, the view-invariant (shared) and view-specific (private) representations are fused to generate the complementary multi-view representation $\boldsymbol{h}^{l+1}$. The fusion process is defined as follows:
\begin{equation}
    {\boldsymbol{ h}}^{l+1}  =  \underbrace{\operatorname{Sum}_{v \in \mathcal V}(\boldsymbol{\tilde h}^{\,l}_{v,s})}_{\text{\textbf{Shared Branches}}} +  \underbrace{\operatorname{Concat}_{v \in \mathcal V}(\boldsymbol{\tilde h}^{\,l}_{v,p}) \mathbf{W}^{l}_{p}}_{\text{\textbf{Private Branches}}},
\label{eq:fusion}
\end{equation}
where $\mathbf{W}^{l}_{p}\in\mathbb{R}^{4D\times D}$ denotes the private projection matrix. Notably, all disentangled transformations are implemented using lightweight projection matrices, which incur lower overhead than MLP layers and preserve the efficiency advantages of the Mamba architecture.

\subsection{Training and Inference}
\subsubsection{\textbf{Training Objective}}
While the proposed disentangled Mamba backbone structurally separates shared and private branches at the architectural level, disentanglement along feature pathways remains challenging during training. Without explicit constraints, shared information may still leak into private branches, causing representational crosstalk and undermining the disentanglement objective. To address this issue, we incorporate a disentanglement regularization term to explicitly regulate the separation between shared and private components. Specifically, a lightweight cross-covariance term is introduced to reduce statistical dependency between representation components~\cite{zbontar2021barlow}. This loss explicitly enforces the decoupling of view-invariant and view-specific representations, thereby stabilizing the disentangled fusion process.
\begin{equation}
\mathcal{L}_{\mathrm{DIS}}
=
\frac{1}{L\,|\mathcal V|}
\sum_{l=0}^{L-1}
\sum_{v \in \mathcal V}
\operatorname{CrossCov}
\left(
\operatorname{StopGrad}(\boldsymbol{\tilde H}^{\,l}_{v,s}),
\boldsymbol{\tilde H}^{\,l}_{v,p}
\right),
\end{equation}
\begin{equation}
\operatorname{CrossCov}(\boldsymbol H_1, \boldsymbol H_2)
=
\left\|
\frac{1}{B-1}
\hat{\boldsymbol H}_1^{\top}
\hat{\boldsymbol H}_2
\right\|_{Fro}^2 ,
\end{equation}
\begin{equation}
\hat{\boldsymbol H}
=
\frac{
\boldsymbol H - \mathbb{E}[\boldsymbol H]
}{
\sqrt{\mathrm{Var}(\boldsymbol H)} + \varepsilon
},
\end{equation}
where $\boldsymbol{\tilde H}^{\,l}_{v,s}$ and $\boldsymbol{\tilde H}^{\,l}_{v,p}$ denote batch-level shared and private embeddings. $L$, $|\mathcal V|$, and $B$ denote the numbers of Mamba layers, scanning views, and batch samples, respectively. $\operatorname{StopGrad}(\cdot)$ blocks gradients to shared representations, and $\varepsilon$ ensures numerical stability.

Reconstruction loss plays a central role under the zero-positive training setting of \task. The final Mamba layer outputs the encoded representation $\boldsymbol{h}^{L}$. Following the standard autoencoder paradigm, we apply a linear projection~\cite{tan2025meet} to map $\boldsymbol{h}^{L}$ back to the input space, yielding the reconstructed output $\boldsymbol{\hat{x}}$. To promote accurate reconstruction of normal traffic patterns, we adopt a cosine similarity loss $\mathcal{L}_{\text{REC}}$ between the input traffic and its reconstruction. Thus, the overall training objective combines the reconstruction loss and the disentangled constraint:
\begin{equation}
\mathcal{L}_{\text{REC}}= \sum_{i=1}^B(1- \frac{\boldsymbol {x^T_i} \ \boldsymbol{\hat x}_i}{\| \boldsymbol x_i \|\| \boldsymbol{\hat x}_i \|}),
\end{equation}
\begin{equation}
\mathcal{L}= \mathcal{L}_{\text{REC}}+ \mathcal{L}_{\text{DIS}}.
\label{eq:total_loss}
\end{equation}
More model and training details are provided in Appendix~\ref{sec:appendix_algorithm}.

\subsubsection{\textbf{Anomaly Scoring}}
Under the zero-positive training paradigm, \method~learns to accurately reconstruct normal traffic patterns. When anomalous traffic is encountered, the input deviates from the learned representations, leading to larger reconstruction errors. Accordingly, we compute the anomaly score using the cosine dissimilarity between the original input $\boldsymbol{x}$ and its reconstruction $\boldsymbol{\hat{x}}$, where higher scores indicate a higher likelihood of anomalous behavior for the test sample:
\begin{equation}
s= 1- \frac{\boldsymbol{x}^{T} \ \boldsymbol{\hat x}}{\| \boldsymbol{x} \|\| \boldsymbol{\hat x} \|}.
\end{equation}

\begin{table*}[t]
	\centering
	\setlength\tabcolsep{4pt}
	\belowrulesep=0.7pt
    \caption{\upshape Performance comparisons (\%) on three public datasets. The best results are in \textbf{bold} font and the second \underline{underlined}.}\label{tab:benchmark}
    \vspace{-0.3cm}
	\begin{tabular}{cccccccccc} 
	\toprule
	\multirow{2}{*}{\textbf{Methods}} & \multicolumn{3}{c}{\textbf{CIC-IoT2023}} & \multicolumn{3}{c}{\textbf{DoHBrw2020}} & \multicolumn{3}{c}{\textbf{ISCX-Tor2016}}  \\
	\cmidrule(lr){2-4} \cmidrule(lr){5-7} \cmidrule(lr){8-10}
	& \textbf{AUC(\%)}   & \textbf{ACC(\%)}  & \textbf{F1(\%)}            & \textbf{AUC(\%)}   & \textbf{ACC(\%)}  & \textbf{F1(\%)}              & \textbf{AUC(\%)}   & \textbf{ACC(\%)}  & \textbf{F1(\%)}             \\ 
	\toprule	
DAGMM               & 66.40 & 61.38 & 67.22           & 77.41 & 75.18 & 75.07          & 73.28 & 74.30 & 73.82            \\
ARCADE              & 72.09 & 77.32 & 74.94           & 76.79 & 79.91 & 80.07         & 92.35 & 90.15 & 90.50            \\
MFR                 & 82.68 &  83.03 & 81.56           & 70.51 & 70.23 & 73.44          &  93.68 & 91.74 & \underline{91.94}            \\
UnDiff              &  83.25 & 82.69 & 81.93           & 81.92 & 79.93 & 80.10   &  94.01 & \underline{91.86} & 91.61           \\
CountMamba             & 75.38  &  68.37 &  74.10         & 80.45  & 78.68  &  81.89         &   92.90  &  89.32  &  89.80           \\
ConMD             & \underline{83.76}  &  \underline{83.11}  &  81.80      & \underline{83.44}  & \underline{80.14}  &  \underline{82.09}   & \underline{94.35} &  91.79 &  91.23        \\
\midrule
Anomaly Transformer & 64.63 & 67.15 & 71.86           & 78.86 & 78.55 & 78.87          & 92.27 & 90.19 & 90.52            \\
TSLANet             & 81.64 & 76.62 & 80.04           & 77.04 & 68.93 & 73.10          & 93.63 & 90.97 & 90.71            \\
iTransformer        & 81.32 & 82.07 &  \underline{81.95}           &  82.44 & 78.66 & 78.26          & 82.65 & 81.34 & 83.97            \\
SMamba             & 82.67 & 81.35  & 80.07           &  82.92 & 78.16  &  80.63         & 94.00  & 91.45  &  91.42           \\
MambaAD             & 82.29 & 77.70  & 80.02           &  82.79 & 79.36  &  81.85         & 92.54  & 85.21  &  85.48           \\
\midrule
\textbf{\method}             & \textbf{86.81} & \textbf{84.66} & \textbf{83.82}           &   \textbf{87.34}    &    \textbf{83.27}    &    \textbf{83.86}            & \textbf{96.40} & \textbf{92.95} & \textbf{93.00}           \\
	\bottomrule
	\end{tabular}
\label{Table:MainResults}
\end{table*}
\section{Experiments}
\label{sec:Experiments}
\subsection{Experimental Setup}
\noindent\textbf{Datasets.} We evaluate on three network traffic anomaly detection datasets featuring encrypted payload traffic and diverse network topologies. (1) \emph{CIC-IoT2023}: a large-scale intrusion detection dataset covering seven attack categories under Internet of Things topologies~\cite{neto2023ciciot2023}. (2) \emph{DoHBrw2020}: a malicious HTTPS traffic dataset centered on DNS-over-HTTPS behaviors~\cite{montazerishatoori2020detection}. (3) \emph{ISCX-Tor2016}: an anonymous Tor traffic dataset that captures traffic patterns beyond payload encryption, enabling evaluation under anonymity-network scenarios~\cite{lashkari2017characterization}. Following prior work~\cite{lian2025mfr}, we randomly sample 10,000 normal flows for training and construct a balanced test set with 5,000 normal and 5,000 anomalous flows.

\noindent\textbf{Evaluation Metrics.} We assess model performance using three standard metrics: Area Under the Curve (AUC), Accuracy (ACC), and Macro F1-Score (F1), following recent studies~\cite{zheng2023semi,lian2025facing}. We determine the anomaly-score threshold by maximizing the F1-score, which balances precision and recall.

\noindent\textbf{Baselines.} Considering the sequential nature of network traffic, we compare \method~with eleven competitive baseline models, which can be divided into two groups: (1) \textit{Network traffic} anomaly detection methods: DAGMM ~\cite{zong2018deep}, ARCADE~\cite{lunardi2023arcade}, MFR~\cite{lian2025mfr}, UnDiff~\cite{lian2025facing}, ConMD~\cite{lian2026contextual}, and CountMamba~\cite{deng2025countmamba}. (2) \textit{Time-series} anomaly detection methods: Anomaly Transformer~\cite{xu2022anomaly}, TSLANet~\cite{eldele2024tslanet}, iTransformer~\cite{liu2024itransformer}, SMamba~\cite{Wang2025ismamba}, and MambaAD~\cite{qin2025mambaad}.

\noindent\textbf{Testbeds.} All experiments are conducted on a computational platform equipped with an Intel i5-14600KF 3.50 GHz CPU, an NVIDIA GeForce 4090 GPU, and Ubuntu 20.04. We implement all neural network models using the PyTorch framework.

\noindent\textbf{Implementation Details.} For the model setup, we set the number of disentangled Mamba blocks $L$ to 3, the token embedding dimension $D$ to 32, and the low-rank dimension $R$ of private branches to 8. For traffic representation, we set $F=4$ packets per flow and $P=600$ bytes per packet. The same hyperparameter settings are used across the three datasets. We use the Adam optimizer with a learning rate of $1\times10^{-4}$ and a batch size of 128. Each model is trained for up to 50 epochs, with early stopping applied to prevent overfitting. We perform five independent runs with different random seeds and report the mean performance across runs.

\begin{figure}[t]
\centerline{\includegraphics[width=0.85\linewidth]{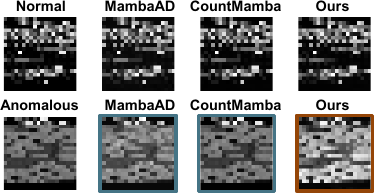}} 
\vspace{-0.3cm}
\caption{Reconstruction comparison on representative cases.}
\label{fig:recon}
\vspace{-0.3cm}
\end{figure}

\subsection{Main Results}
We compare \method~with eleven strong baselines to evaluate its effectiveness. The quantitative results are reported in Table~\ref{Table:MainResults}. Based on these results, we make the following observations.

\noindent\textbf{(O1)}: \method~consistently outperforms all baseline methods across different datasets. In particular, \method~achieves AUC improvements of 3.1\% on CIC-IoT2023 and 3.9\% on DoHBrw2020. These improvements primarily arise from the proposed disentangled multi-view Mamba scanning structure. By adopting a disentangle-first design, \method~effectively prevents the accumulation of view-invariant redundancy as more scanning views are incorporated. This design preserves complementary multi-view contextual cues, enhancing sensitivity to subtle multi-view anomalous patterns in network traffic~\cite{lian2026contextual}. As shown in~\figurename~\ref{fig:recon}, \method~produces more deviated reconstructions for anomalous traffic while maintaining high-fidelity reconstructions for normal samples, thereby facilitating more reliable anomaly detection.

\noindent\textbf{(O2)}: Existing \task~methods, such as MFR and ConMD, mainly improve detection performance through enhanced feature extraction, but are predominantly built on CNN-based architectures. Although CNNs exhibit strong local modeling capability, they are less effective at capturing long-range dependencies and multi-view characteristics inherent in network traffic, which limits their detection performance. While the recent CountMamba introduces Mamba into \task, it is restricted to unidirectional scanning. This design introduces causal bias, constraining its ability to model multi-view traffic patterns and resulting in incomplete contextual modeling.

\noindent\textbf{(O3)}: Time-series-based approaches, including Anomaly Transformer, TSLANet, and iTransformer, generally underperform compared to specialized \task~methods. This is because these models primarily focus on temporal trends and fluctuations, which are less informative for payload-encrypted network traffic. Moreover, recent Mamba-based time-series models adopt multi-view scanning but typically overlook redundancy accumulation caused by naive fusion strategies. This leads to homogenized and degraded feature representations across views. As illustrated in~\figurename~\ref{fig:recon}, although these methods reconstruct normal traffic with high fidelity, they tend to overgeneralize anomalous samples~\cite{lian2026contextual}, producing similarly accurate reconstructions. This observation reflects the limited sensitivity of their homogenized representations to subtle multi-view anomalous patterns in network traffic.

\begin{table}[t]
\setlength\tabcolsep{3pt}
\belowrulesep=0.8pt
\centering
\caption{\upshape Ablation results on main components (AUC).}
\vspace{-0.3cm}
\begin{tabular}{ccccc}
\toprule
 \textbf{Variant} & \begin{tabular}[c]{@{}c@{}}\textbf{IoT2023}\end{tabular} & \begin{tabular}[c]{@{}c@{}}\textbf{DoHBrw2020}\end{tabular} & \begin{tabular}[c]{@{}c@{}}\textbf{Tor2016}\end{tabular}  \\
\toprule
w/ Unidirectional   &   82.14     &   83.84    &    93.21     \\
w/ Summation   &    77.30    &   84.11    &    93.52     \\
w/o Disentangle   &   79.02     &  83.02     &     93.89     \\
\midrule
w/o Shared  Branches &    77.84    &   85.51   &   95.79      \\
w/o Private Branches  &   75.15     &  83.43     &    94.36     \\
w/o Disentanglement Loss   &   84.26     &   85.76    &  94.83      \\
\midrule
w/o Low-Rank Projection   &   84.40     &   85.12    &  94.38      \\
w/o Stop-Gradient   &   85.77     &   87.01    &  95.98      \\
\midrule
\textbf{\method}     & \textbf{86.81}       & \textbf{87.34}      & \textbf{96.40}       \\
\bottomrule
\end{tabular}
\label{Table:ablation}
\vspace{-0.25cm}
\end{table}

\subsection{Ablation Study}
We evaluate our disentangled Mamba variants by removing key components. The results are reported in Table~\ref{Table:ablation}, and the main observations can be summarized in three aspects.

\noindent \textbf{Effect of Multi-View Scanning:} Replacing the proposed multi-view scanning with a simple unidirectional strategy (w/ Unidirectional) leads to noticeable performance degradation, indicating that unidirectional scanning is biased. It is insufficient for capturing the complementary multi-view characteristics of network traffic. In addition, when adopting the commonly used feature-level summation for multi-view fusion (w/ Summation), the performance becomes comparable to the unidirectional setting, and even worse on the CIC-IoT2023 dataset. This result validates the existence and negative impact of multi-view redundancy accumulation. We further remove the entire disentanglement design while retaining the proposed projection modules (w/o Disentangle). The observed degradation demonstrates that the performance improvements stem from the proposed disentangled scanning mechanism rather than from the introduction of additional network layers.

\noindent \textbf{Effect of Disentanglement Architecture:} Removing either the shared branches (w/o Shared Branches) or the private branches (w/o Private Branches) results in substantial performance drops. This confirms that both view-invariant and view-specific components are indispensable for representation learning, with discriminative view-specific information playing a more critical role. Furthermore, removing the disentanglement constraint loss (w/o Disentanglement Loss) also leads to a clear performance degradation, indicating that structural disentanglement alone is insufficient without explicit optimization constraints during training.

\noindent \textbf{Effect of Private-Branch Constraints:} Removing the low-rank projection (w/o Low-Rank Projection) causes consistent performance drops, indicating that structural capacity restriction helps prevent invariant information from leaking into private representations. Removing the stop-gradient operation (w/o Stop-Gradient) also degrades performance, showing that gradient isolation from the shared branch stabilizes private representation extraction. The redundancy analysis in Appendix~\ref{sec:appendix_ablation_visualize} provides additional evidence that both constraints reduce inter-view redundancy, confirming their role in mitigating redundancy accumulation and improving view-specific representation learning.

\begin{figure}[t]
\centerline{\includegraphics[width=0.9 \linewidth]{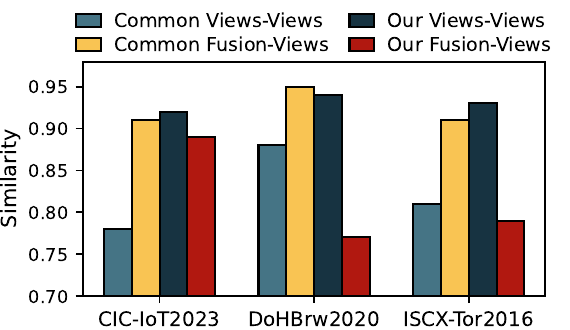}}
\vspace{-0.3cm}
\caption{CKA similarity before and after disentanglement.}
\label{fig:CKA-our}
\vspace{-0.0cm}
\end{figure}
\begin{figure}[t]
\centerline{\includegraphics[width=0.9 \linewidth]{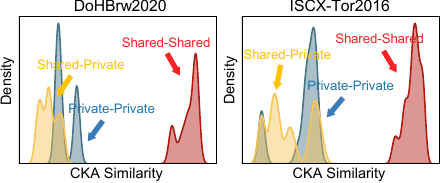}}
\vspace{-0.3cm}
\caption{Density distributions of CKA similarity for inter-private, private--shared, and inter-shared representations.}
\label{fig:CKA-density}
\vspace{-0.25cm}
\end{figure}

\subsection{Redundancy Reduction Analysis}
\noindent\textbf{Disentanglement Analysis}: \figurename~\ref{fig:CKA-our} quantitatively demonstrates the redundancy reduction effect of \method. Consistent with the phenomenon observed in~\figurename~\ref{fig:redundancy1}, common feature-level summation repeatedly reinforces view-invariant information in the fused representations, as reflected by increased CKA similarity after fusion (Common Fusion--Views) relative to before fusion (Common Views--Views). In contrast, under our disentangled design, the CKA similarity between the fused representation and the scanning views (Our Fusion--Views) is noticeably lower than that between different scanning views (Our Views--Views). This observation indicates that our framework effectively integrates complementary view-specific information rather than repeatedly amplifying view-invariant components, thereby alleviating multi-view representation homogenization and redundancy accumulation.

Moreover, \figurename~\ref{fig:CKA-density} further verifies the effectiveness of the proposed disentanglement mechanism. The similarities among shared branches (Shared–Shared) exhibit a highly concentrated distribution, with non-overlapping similarity distributions relative to other branch groups, indicating that view-invariant information is successfully isolated and consistently aggregated. In contrast, both inter-private (Private–Private) and shared–private (Shared–Private) similarities present low-value distributions. These results indicate that shared and private representations are effectively separated, while each private branch consistently captures view-specific information. Overall, \method~achieves effective feature disentanglement and reduces redundancy accumulation.

\begin{figure}[t]
\centering
\begin{minipage}{0.5\columnwidth}
  \centering
  \includegraphics[width=\linewidth]{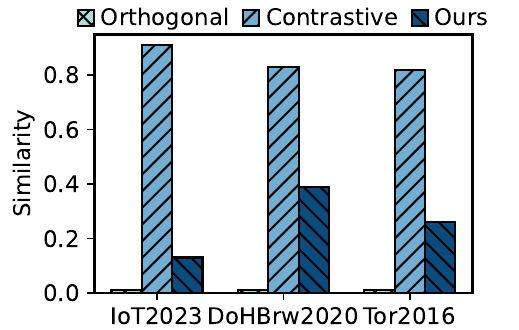}
    \vspace{-0.6cm}
  \caption*{(a) CKA similarity}
\end{minipage}\hfill
\begin{minipage}{0.5\columnwidth}
  \centering
  \includegraphics[width=\linewidth]{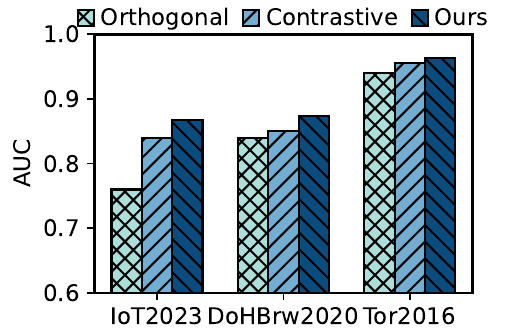}
    \vspace{-0.6cm}
  \caption*{(b) AUC performance}
\end{minipage}
\vspace{-0.3cm}
\caption{Comparison with other disentanglement methods.}
\label{fig:cka_compare}
\vspace{-0.0cm}
\end{figure}

\noindent\textbf{Comparison with Other Disentanglement Methods}: We further compare \method~with representative feature disentanglement strategies, including orthogonal projection~\cite{jaehoon2023orthogonality} and contrastive learning~\cite{stefan2023towards}. The CKA similarities between shared and private branches, along with the fusion AUC performance, are shown in~\figurename~\ref{fig:cka_compare}(a) and (b). Orthogonality-based methods attempt to eliminate redundancy by enforcing strict orthogonality constraints between representations. Although this strategy achieves strong feature disentanglement, the overly rigid constraint disrupts inter-branch interactions. As a result, fusion effectiveness is degraded, leading to inferior detection performance. In contrast, contrastive-based methods construct positive and negative pairs to guide representation alignment and disentanglement. However, experimental results show that this strategy fails to remove redundancy in the Mamba architecture. Our \method~achieves a better balance between disentanglement quality and fusion effectiveness, leading to superior anomaly detection performance.

\begin{figure}[t]
\centering
\begin{minipage}{0.50\columnwidth}
  \centering
  \includegraphics[width=\linewidth]{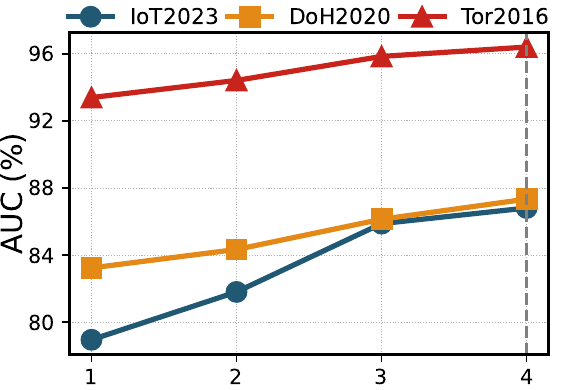}
  \vspace{-0.75cm}
  \caption*{(a) Scanning View Number $|\mathcal V|$}
\end{minipage}\hfill
\begin{minipage}{0.50\columnwidth}
  \centering
  \includegraphics[width=\linewidth]{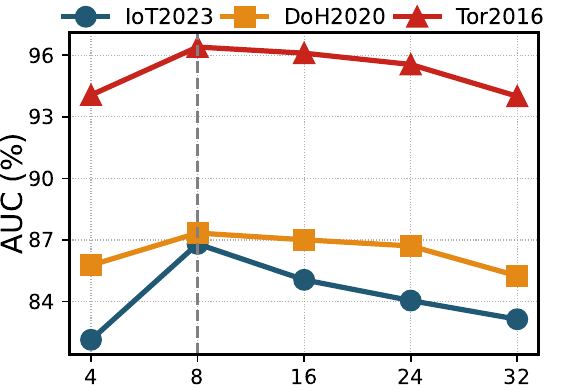}
  \vspace{-0.75cm}
  \caption*{(b) Low-Rank Dimension $R$}
\end{minipage}
\vspace{-0.3cm}
\caption{Hyperparameter sensitivity analysis across datasets.}
\label{fig:Hyper}
\vspace{-0.25cm}
\end{figure}

\subsection{Hyperparameter Analysis}
\figurename~\ref{fig:Hyper} presents the sensitivity analysis of two key hyperparameters in~\method. Additional results are reported in Appendix~\ref{sec:appendix_hyperparameter}. The main observations are summarized as follows.

\noindent\textbf{(O1)}: The number of scanning views $|\mathcal V|$ controls the amount of encoded multi-view information. Results show that detection performance consistently improves as $|\mathcal V|$ increases, indicating that \method~benefits from richer multi-view representations. With the proposed disentangle-then-fuse design, \method~can effectively integrate complementary view cues, enabling more accurate detection of subtle traffic anomalies.

\noindent\textbf{(O2)}: The low-rank dimension $R$ regulates information throughput within private branches and is a critical structural hyperparameter. An excessively large $R$ leads to view-invariant information leakage from shared branches, weakening the disentanglement effect. Conversely, an overly small $R$ severely restricts the throughput capacity of private branches, resulting in the loss of view-specific information and eventual disentanglement collapse.

\subsection{Overhead Analysis}
\begin{figure}[t]
\centerline{\includegraphics[width=0.9\linewidth]{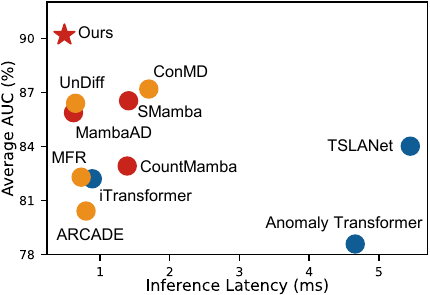}}
\caption{Inference overhead and performance comparison.}
\label{fig:overhead}
\vspace{-0.3cm}
\end{figure}

Inference overhead is a critical factor for practical deployment of \task~systems. \figurename~\ref{fig:overhead} compares detection performance (average AUC across three datasets) with per-sample inference latency for different model families. CNN-based methods (orange markers) exhibit linear computational complexity $\mathcal{O}((F \cdot P)D)$~\cite{gao2025ssd}, but rely heavily on feature extraction engineering, which introduces non-negligible preprocessing overhead. Transformer-based approaches (blue markers) suffer from quadratic complexity $\mathcal{O}((F \cdot P)^2 D)$, leading to substantial computational costs, especially for traffic sequences. In contrast, Mamba-based architectures (red markers) achieve linear complexity $\mathcal{O}((F \cdot P)D)$, offering significantly improved efficiency. Building upon this efficient backbone, our disentangled Mamba framework introduces only lightweight projection modules, incurring negligible additional overhead. As a result, \method~achieves superior detection performance while maintaining millisecond-level inference latency.
\section{Related Work}
\label{sec:Related Work}
\subsection{Network Traffic Anomaly Detection}
Network Traffic Anomaly Detection (NTAD) aims to identify previously unseen anomalous traffic using only normal samples during training~\cite{zhang2024aoc, dang2025semi}. Most methods follow a reconstruction-based paradigm, where reconstruction errors are used as anomaly indicators. They also rely heavily on feature extraction to improve detection performance. Representative methods such as MFAD~\cite{zheng2023semi} and MFR~\cite{lian2025mfr} exploit low-pass filtering to emphasize low-frequency texture-level patterns for better normal traffic modeling. ARCADE~\cite{lunardi2023arcade} incorporates adversarial training into autoencoder architectures to enhance reconstruction quality. Trident~\cite{zhao2024trident} adopts a U-Net-based autoencoder to preserve fine-grained structural characteristics of network traffic. unFlowS~\cite{yang2025unflows} projects reconstruction errors into spectral spaces to expose anomalous patterns. UnDiff~\cite{lian2025facing} introduces the first distribution-aware anomaly scoring method, using uncertainty to evaluate anomalous deviations beyond deterministic reconstruction errors. FreeUp~\cite{lian2026decompose} proposes a frequency-decoupled reconstruction framework for modeling encrypted ``full-frequency'' network traffic.

Despite these advances, most NTAD methods are built upon CNN-based architectures that primarily focus on local pattern modeling. They fail to capture long-range contextual dependencies and multi-view characteristics. To alleviate this limitation, ConMD~\cite{lian2026contextual} introduces a knowledge distillation–based feature extraction scheme to learn discriminative representations and adopts a CNN--Transformer hybrid architecture to enhance contextual modeling. Given the sequential nature of network traffic, time-series anomaly detection methods are also considered as competitive baselines. These methods predominantly adopt Transformer-based architectures to model sequential trends and fluctuations~\cite{xu2022anomaly, eldele2024tslanet, liu2024itransformer}. However, Transformers inherently suffer from quadratic computational complexity. In contrast, our \method~is built upon State Space Models and captures sequential traffic dependencies with linear-time complexity. It further employs a lightweight disentangled multi-view scanning design, achieving superior detection performance with high inference efficiency.

\subsection{Multi-View Scanning in Mamba}
Mamba is built upon selective state-space sequence models and enables contextual representation learning with linear computational complexity. It has demonstrated a favorable efficiency–effectiveness trade-off in domains such as computer vision and time-series analysis~\cite{zhang2025wave,qin2025mambaad}. The core capability of Mamba lies in its multi-view scanning mechanism, which is typically customized for different application domains~\cite{huang2024localmamba, gao2025ssd, Wang2025ismamba}. For example, in computer vision, representative scanning strategies such as Slanted Scan, Zigzag Scan, and Hilbert Scan arrange sequences in specific geometric orders~\cite{qu2024survey, liu2025vision}. By fusing representations from multiple scanning pathways, Mamba aggregates complementary directional information and constructs globally coherent contextual representations~\cite{liu2025vision}. However, recent studies on network traffic analysis report that directly adopting multi-view Mamba scanning can lead to performance degradation~\cite{wang2024netmamba}. Consequently, several networking-oriented works resort to simplified unidirectional Mamba architectures~\cite{deng2025countmamba,xia2025mamba4net}. While this alleviates degradation issues, it also sacrifices critical multi-view contextual awareness required for anomaly detection in network traffic~\cite{lian2026contextual}. Notably, some studies in the computer vision domain have also reported representation redundancy induced by multi-view scanning and its adverse and detrimental impact on performance~\cite{guo2024mambair,shi2024multi,xiao2025boosting}. Nevertheless, these works remain at a largely intuitive level, lacking systematic mechanism analysis and quantitative evidence.

In contrast, our work conducts comprehensive quantitative and qualitative analyses of scanning representations. We systematically identify a redundancy accumulation phenomenon and a multi-view degradation problem in network traffic scenarios. To address these, we propose a novel disentangled Mamba framework that explicitly separates view-invariant and view-specific information. By jointly balancing invariant and view-specific components, \method~mitigates redundancy-induced representation degradation while preserving multi-view cues, thereby unlocking Mamba’s representational capacity for NTAD.

\subsection{Disentangled Representation Learning}
Disentangled representation learning aims to decompose complex representations into semantically distinct factors, thereby improving interpretability, robustness, and generalization~\cite{cheng2025disentangling}. A common strategy is to separate representations into shared and private components: shared components capture invariant information across views or domains, whereas private components preserve view-specific or domain-specific characteristics~\cite{liang2025multi}. Representative studies achieve this separation through four types of strategies: orthogonality-based~\cite{jaehoon2023orthogonality}, contrastive-learning-based~\cite{stefan2023towards}, causality-inspired~\cite{cheng2023enhancing}, and information-theoretic methods~\cite{liang2025multi}. Specifically, they reduce representation entanglement through orthogonal subspace constraints, contrastive factor separation, causal invariance modeling, or mutual-information regularization. While shared-private decomposition appears suitable for addressing redundancy in multi-view Mamba scanning, directly applying existing disentanglement methods remains non-trivial. This is because they mainly target generic factor separation, while overlooking the dominant view-invariant information shared across scanning views and amplified during fusion. Directly applying these methods may either lead to over-separation that disrupts semantic consistency or fail to sufficiently separate shared and private components in Mamba representations. In contrast, \method~constructs an efficient disentangled framework through lightweight projection modules and structural constraints. It further reformulates multi-view scanning in Mamba as a disentangle-then-fuse process, facilitating more effective representation learning.
\section{Conclusions}\label{sec:Conclusions}
In this paper, we identify, for the first time, a critical redundancy accumulation phenomenon in applying Mamba to NTAD, revealing a previously overlooked structural deficiency in multi-view Mamba scanning. Specifically, naive multi-view extensions repeatedly amplify view-invariant information, leading to multi-view representation degradation. To address this deficiency, we propose \method, a disentangled multi-view Mamba framework that reformulates multi-view scanning as a disentangle-then-fuse process. By explicitly separating view-invariant and view-specific representations before fusion, \method~mitigates redundancy accumulation while preserving complementary multi-view cues, thereby improving sensitivity to subtle anomalous traffic patterns. Extensive experiments demonstrate the effectiveness of our \method, achieving consistently superior detection performance with millisecond-level inference efficiency. Overall, this work establishes a disentangle-then-fuse paradigm for multi-view scanning in Mamba, opening new opportunities for Mamba architecture design. In future work, we will explore extending this framework beyond NTAD to broader application scenarios, such as computer vision, time-series analysis, and language modeling.

\begin{acks}
This work was supported by the National Natural Science Foundation of China (Grant No. 62572097).
\end{acks}


\clearpage


\bibliographystyle{ACM-Reference-Format}
\balance
\bibliography{sample-base}

\setcounter{secnumdepth}{2}
\appendix
\label{sec:Appendix}

\section{Traffic Data Preprocessing}\label{sec:data preprocessing}
The basic unit of network traffic is a flow, which consists of a sequence of ordered packets. Typically, a traffic sample is defined as a flow whose packets share the same 5-tuple, including the transport-layer protocol, source IP address, destination IP address, source port, and destination port~\cite{lian2025mfr}. Following this definition, we first segment raw traffic into flows based on the 5-tuple, and then preprocess each flow through address anonymization, header-length normalization, and sequence padding/truncation.

\vspace{0.1cm}
Specifically, we remove the Ethernet header and set IP addresses to zero to prevent the model from overfitting to address-specific patterns. Since the header lengths of User Datagram Protocol (UDP) and Transmission Control Protocol (TCP) segments are inconsistent, we pad the UDP header with zero bytes (0x00) to match the TCP header length, thereby alleviating the negative impact of transport-layer length discrepancies. Finally, following prior work~\cite{lian2025mfr}, we apply standard truncation and padding operations to unify flow and packet lengths. Based on our hyperparameter study, the flow length and packet length are set to $F=4$ and $P=600$, respectively.

\begin{figure*}[t]
\centering
\begin{minipage}{1\columnwidth}
  \centering
  \includegraphics[width=0.98\linewidth]{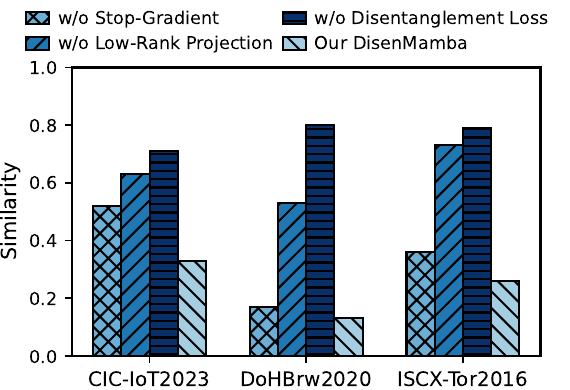}\vspace{0.1cm}\\
    {\small\textbf{(a) CKA Similarity between Shared and Private Branches}}
\end{minipage}\hfill
\begin{minipage}{1\columnwidth}
  \centering
  \includegraphics[width=0.98\linewidth]{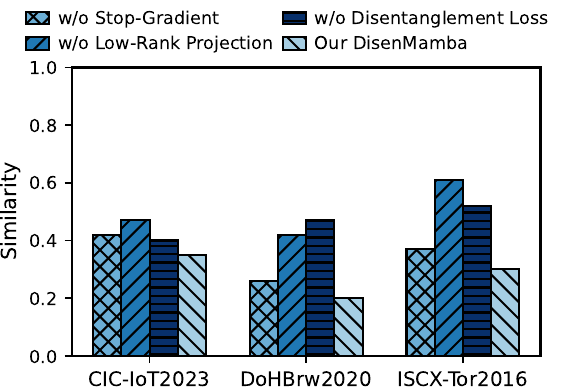}\vspace{0.1cm}\\
    {\small\textbf{(b) CKA Similarity between Inter-Private Branches}}
\end{minipage}
\vspace{-0.2cm}
\caption{CKA-based inter-branch redundancy statistics under different ablation variants.}
\vspace{-0.05cm}
\label{fig:appendix_cka-ablation}
\end{figure*}

\section{Algorithm Description}\label{sec:appendix_algorithm}
Algorithm~\ref{training} presents the \method~training pipeline, including the disentangle-then-fuse process.

\begin{algorithm}[htb]
\renewcommand{\algorithmicrequire}{\textbf{Input:}}
\renewcommand{\algorithmicensure}{\textbf{Output:}}
\caption{\method~Training Pipeline}\label{training}
\begin{algorithmic}[1]

\REQUIRE Mini-batch traffic samples $\boldsymbol{X} \in \mathbb{R}^{B \times F \times P}$, where $B$ is the batch size, $F$ is the number of packets in each flow, and $P$ is the number of bytes in each packet

\STATE \underline{$\textbf{\textit{Traffic Embedding}}$}
\STATE $\boldsymbol{H}^{0} = \operatorname{Conv1D}(\boldsymbol{X}) + \operatorname{Sine-Cosine}(\boldsymbol{X})$, $\boldsymbol{H}^{0} \in \mathbb{R}^{B \times F \times P \times D}$

\STATE \underline{$\textbf{\textit{Disentangled Multi-View Mamba Encoder}}$}
\FOR{$l = 0$ to $L-1$}

    \STATE \textit{/*Multi-View Scanning*/}
    \vspace{0.04cm}
    \FOR{each $v\in\mathcal{V} = \{\vec{\mathrm{F}}, \vec{\mathrm{P}}, \lvec{\mathrm{F}}, \lvec{\mathrm{P}}\}$}

    \STATE Rearrange multi-view:
    $\boldsymbol{H}^{\,l}_{v}$

    \STATE Construct representation:
    $\boldsymbol{\tilde H}^{\,l}_{v} = \operatorname{MambaBlock}\!\left(\boldsymbol{H}^{\,l}_{v} \right)$
    
    \ENDFOR
    
    \STATE \textit{/*Shared Branch Extraction*/}
    \vspace{0.08cm}
    
    \STATE $\boldsymbol{\tilde H}^{\,l}_{v,s}
    = \boldsymbol{\tilde H}^{\,l}_{v}\mathbf{W}^{l}_{s}$

    \vspace{0.04cm}
    \STATE \textit{/*Private Branch Extraction*/}
    \vspace{0.06cm}
    
    \STATE $\boldsymbol{\tilde H}^{\,l}_{v,p}
    =
    (\boldsymbol{\tilde H}^{\,l}_{v}
    -
    \operatorname{StopGrad}(\boldsymbol{\tilde H}^{\,l}_{v,s}))
    \mathbf{W}^{l}_{v,p,1}\mathbf{W}^{l}_{v,p,2}$

    \vspace{0.04cm}
    \STATE \textit{/*Disentangled Multi-View Fusion*/}
    \vspace{0.04cm}
    
    \STATE Shared aggregation:
    $\boldsymbol{\tilde H}^{\,l}_{s}
    =
    \sum_{v \in \mathcal V}
    \boldsymbol{\tilde H}^{\,l}_{v,s}$

    \STATE Private aggregation:
    $\boldsymbol{\tilde H}^{\,l}_{p}
    =
    \operatorname{Concat}(\{\boldsymbol{\tilde H}^{\,l}_{v,p}\}_{v \in \mathcal V})
    \mathbf{W}^{l}_{p}$

    \STATE Final fusion:
    $\boldsymbol{\tilde H}^{l}
    =
    \boldsymbol{\tilde H}^{\,l}_{s}
    +
    \boldsymbol{\tilde H}^{\,l}_{p}$

    \vspace{0.04cm}
    \STATE \textit{/*Feed-Forward Network Layer*/}
    \STATE $\boldsymbol{H}^{l+1} = \operatorname {Conv1D}( \operatorname{ReLU}(\operatorname {Conv1D}(\operatorname{LN} (\boldsymbol{\tilde H}^{l}))))$

\ENDFOR

\STATE \underline{$\textbf{\textit{Decoder Projection}}$}
\STATE $\boldsymbol{\hat{X}}
=
\operatorname{Linear}(\boldsymbol{H}^{L})$

\STATE \underline{$\textbf{\textit{Loss Computation}}$}

\STATE Reconstruction loss:
\[
\mathcal{L}_{\mathrm{REC}}
=
1 -
\frac{\boldsymbol{X}^{\top}\boldsymbol{\hat{X}}}
{\|\boldsymbol{X}\|\|\boldsymbol{\hat{X}}\|}
\]

\STATE Disentanglement loss:
\[
\mathcal{L}_{\mathrm{DIS}}
=
\frac{1}{L|\mathcal V|}
\sum_{l=0}^{L-1}
\sum_{v \in \mathcal V}
\operatorname{CrossCov}
\big(
\operatorname{StopGrad}(\boldsymbol{\tilde H}^{\,l}_{v,s}),
\boldsymbol{\tilde H}^{\,l}_{v,p}
\big)
\]

\STATE Total loss:
\[
\mathcal{L}
=
\mathcal{L}_{\mathrm{REC}}
+
\mathcal{L}_{\mathrm{DIS}}
\]

\STATE Update parameters via Adam optimizer

\ENSURE Trained \method
\end{algorithmic}
\end{algorithm}

\section{State Space Models}
Mamba is an emerging sequence modeling architecture built upon State Space Models (SSMs) for efficient long-range modeling. 
SSMs are a class of models for sequential data that characterize the relationship between input and output sequences through latent state representations. 
Given a one-dimensional input sequence $\boldsymbol{x} \in \mathbb{R}^O$, an SSM produces a corresponding output sequence $\boldsymbol{y} \in \mathbb{R}^Q$.
In continuous time, an SSM is defined as:
\begin{equation}
\begin{aligned}
\dot{\boldsymbol{h}}(t) &= \boldsymbol{A}\boldsymbol{h}(t) + \boldsymbol{B}\boldsymbol{x}(t), \\
\boldsymbol{y}(t) &= \boldsymbol{C}\boldsymbol{h}(t) + \boldsymbol{D}\boldsymbol{x}(t),
\end{aligned}
\label{eq:ssm_continuous}
\end{equation}
where $\boldsymbol{x}(t) \in \mathbb{R}^O$ and $\boldsymbol{y}(t) \in \mathbb{R}^Q$ denote the input and output at time $t$, respectively; $\boldsymbol{h}(t) \in \mathbb{R}^S$ and $\dot{\boldsymbol{h}}(t) \in \mathbb{R}^S$ represent the hidden state and its time derivative. 
$\boldsymbol{A} \in \mathbb{R}^{S \times S}$, $\boldsymbol{B} \in \mathbb{R}^{S \times O}$, $\boldsymbol{C} \in \mathbb{R}^{Q \times S}$, and $\boldsymbol{D} \in \mathbb{R}^{Q \times O}$ are learnable parameters.

In practice, input sequences are discretely sampled. Following~\cite{gu2022efficiently}, applying the zero-order hold discretization to \equationautorefname~\eqref{eq:ssm_continuous} yields the discrete formulation:
\begin{equation}
\begin{aligned}
\boldsymbol{h}_k &= \bar{\boldsymbol{A}}\boldsymbol{h}_{k-1} + \bar{\boldsymbol{B}}\boldsymbol{x}_k, \\
\boldsymbol{y}_k &= \boldsymbol{C}\boldsymbol{h}_k,
\end{aligned}
\label{eq:ssm_discrete}
\end{equation}
where $\bar{\boldsymbol{A}} = e^{\Delta \boldsymbol{A}}$, 
$\bar{\boldsymbol{B}} = (\Delta \boldsymbol{A})^{-1}(e^{\Delta \boldsymbol{A}} - \boldsymbol{I}) \cdot (\Delta \boldsymbol{B})$, 
and $\Delta$ is the learnable step size.
As shown in \equationautorefname~\eqref{eq:ssm_discrete}, the hidden state $\boldsymbol{h}_k$ is updated recursively based on the current input and the previous state, while the output $\boldsymbol{y}_k$ depends on both the discrete parameters $\boldsymbol{A}$, $\boldsymbol{B}$, $\boldsymbol{C}$ and the input. Since the discretized parameters remain constant across sequence positions, SSMs support parallel training. 
Prior work introduces the High-Order Polynomial Projection Operator (HiPPO) framework~\cite{gu2020hippo} to enhance long-range sequence modeling over continuous input histories.

However, in classical SSMs, the parameters $\boldsymbol{A}$, $\boldsymbol{B}$, and $\boldsymbol{C}$ are time-invariant and input-independent. 
Consequently, the model cannot adaptively assign position-specific importance across the sequence. 
To overcome this limitation, Mamba~\cite{gu2024mamba} introduces input-dependent parameterization to enable selective state-space modeling, significantly enhancing modeling flexibility:
\begin{equation}
\begin{aligned}
{\boldsymbol{B}} &= \boldsymbol W^B \boldsymbol{x}_k,
\end{aligned}
\end{equation}
\begin{equation}
\begin{aligned}
{\boldsymbol{C}} &= \boldsymbol W^C \boldsymbol{x}_k,
\end{aligned}
\end{equation}
\begin{equation}
\begin{aligned}
\Delta &= \tau_\Delta (\boldsymbol W^\Delta \boldsymbol{x}_k),
\end{aligned}
\end{equation}
where $\boldsymbol W^B$, $\boldsymbol W^C$, and $\boldsymbol W^\Delta$ denote input-dependent selective parameter matrices, and $\tau_\Delta (\cdot)$ denotes the softplus activation function. To further address the computational inefficiency of sequential recurrence, Mamba employs a selective scan mechanism that enables efficient parallel computation in linear time while preserving sequence modeling fidelity~\cite{liu2025vision}.

\begin{figure}[t]
\centering
\begin{minipage}{0.50\columnwidth}
  \centering
  \includegraphics[width=\linewidth]{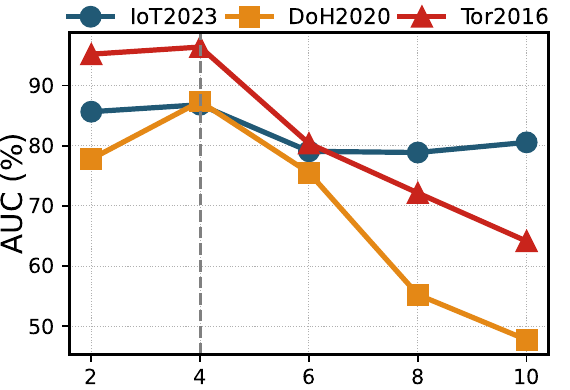}
  \vspace{-0.6cm}
  \caption*{(a) Packet Instance Number $F$}
\end{minipage}\hfill
\begin{minipage}{0.50\columnwidth}
  \centering
  \includegraphics[width=\linewidth]{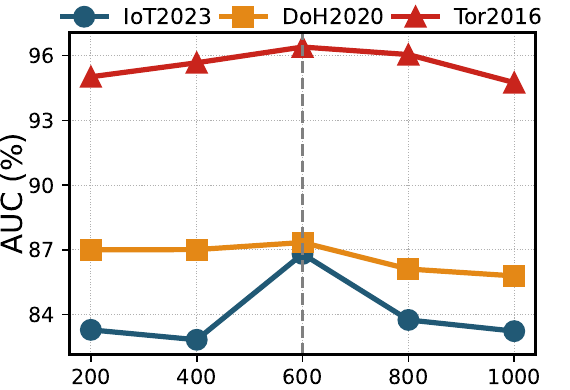}
  \vspace{-0.6cm}
  \caption*{(b) Packet Instance Length $P$}
\end{minipage}
\vspace{-0.3cm}
\caption{Supplementary hyperparameter sensitivity analysis across three datasets.}
\label{fig:apendix_hyper}
\vspace{0.05cm}
\end{figure}

\section{Supplementary Redundancy Analysis}\label{sec:appendix_ablation_visualize}
This section presents a supplementary redundancy analysis of three critical ablation variants in Table~\ref{Table:ablation}: w/o Stop-Gradient, w/o Low-Rank Projection, and w/o Disentanglement Loss. As shown in \figurename~\ref{fig:appendix_cka-ablation}, \figurename~\ref{fig:appendix_cka-ablation}(a) reports the CKA similarity between shared and private branches, indicating shared-private separation, while \figurename~\ref{fig:appendix_cka-ablation}(b) reports the CKA similarity among private branches, reflecting inter-view redundancy within private representations.

\section{Supplementary Hyperparameter Analysis}\label{sec:appendix_hyperparameter}
The data-level representation is crucial for traffic modeling and depends on two key hyperparameters: the number of packets $F$ and the packet byte length $P$. To further analyze their effects, supplementary hyperparameter experiments are conducted, with results reported in detail in \figurename~\ref{fig:apendix_hyper}.

\end{document}